\documentclass{hld2023} 

\usepackage{bm}
\usepackage{cleveref}
\usepackage{float} 

\usepackage[sorting=none]{biblatex}

\bibliography{sample.bib}

\title[Neural Collapse in the Intermediate Hidden Layers of Classification Neural Networks]{Neural Collapse in the Intermediate Hidden Layers of Classification Neural Networks}

\hldauthor{%
\Name{Liam Parker}
\Email{lhparker@princeton.edu}\\
\addr {Princeton University, Princeton, NJ, USA} 
\AND
\Name{Emre Onal} \Email{eonal@princeton.edu}\\
\addr {Princeton University, Princeton, NJ, USA}
\AND
\Name{Anton Stengel} \Email{astengel@princeton.edu}\\
\addr {Princeton University, Princeton, NJ, USA} 
\AND
\Name{Jacob Intrater} \Email{intrater@princeton.edu}\\
\addr {Princeton University, Princeton, NJ, USA} 
}

\begin{document}

\maketitle

\begin{abstract}%
\textit{Neural Collapse} ($\mathcal{NC}$) gives a precise description of the representations of classes in the final hidden layer of classification neural networks. This description provides insights into how these networks learn features and generalize well when trained past zero training error. However, to date, $\mathcal{NC}$ has only been studied in the final layer of these networks. In the present paper, we provide the first comprehensive empirical analysis of the emergence of $\mathcal{NC}$ in the intermediate hidden layers of these classifiers. We examine a variety of network architectures, activations, and datasets, and demonstrate that some degree of $\mathcal{NC}$ emerges in most of the intermediate hidden layers of the network, where the degree of collapse in any given layer is typically positively correlated with the depth of that layer in the neural network. Moreover, we remark that: (1) almost all of the reduction in intra-class variance in the samples occurs in the shallower layers of the networks, (2) the angular separation between class means increases consistently with hidden layer depth, and (3) simple datasets require only the shallower layers of the networks to fully learn them, whereas more difficult ones require the entire network. Ultimately, these results provide granular insights into the structural propagation of features through classification neural networks. 
\end{abstract}

\section{Introduction}
Modern, highly-overparameterized deep neural networks have exceeded human performance on a variety of computer vision tasks \cite{goodfellow2016deep, krizhevsky2017imagenet, lecun2015deep}. However, despite their many successes, it remains unclear how these overparameterized networks converge to solutions which generalize well. In a bid to demystify neural networks' performance, a recent line of inquiry has explored the internally represented `features' of these networks during the Terminal Phase of Training (TPT), i.e. when networks are trained past the point of zero error on the training data \cite{ergen2021revealing, fang2021exploring, zhu2021geometric}. 

The phenomenon of \textit{Neural Collapse} ($\mathcal{NC}$), introduced by Papyan, Han, and Donoho \cite{papyan2020prevalence, han2021neural}, represents one such avenue of inquiry. $\mathcal{NC}$ refers to the emergence of a simple geometry present in neural network classifiers that appears during TPT. Specifically, $\mathcal{NC}$ describes the phenomena by which neural network classifiers converge to learning maximally separated, negligible-variance representations of classes in their last layer activation maps. However, despite the extensive documentation of $\mathcal{NC}$ in the last-layer representations of classification neural networks \cite{mixon2020unconstrained-features, lu2020cross-entropy, tirer2022extended-unconstrained-features, zhou2022different-losses}, there has been no exploration of its presence throughout the intermediate hidden layers of these networks. 

In the present study, we provide a detailed account of the emergence of $\mathcal{NC}$ in the intermediate hidden layers of neural networks across a range of different settings. Specifically, we investigate the impact of varying architectures, datasets, and activation functions on the degree of $\mathcal{NC}$ present in the intermediate layers of classification networks. Our results show that some level of $\mathcal{NC}$ typically occurs in these intermediate layers in all explored settings, where the strength of $\mathcal{NC}$ in a given layer increases as the depth of that layer within the network increases. By examining the presence of $\mathcal{NC}$ not only in the final hidden layer but also the intermediate hidden layers of classification networks, we gain a more nuanced understanding of the mechanisms that drive the behavior of these networks.

\section{Methodology}\label{sec:methodology}
\subsection{Network Architecture and Training}
We examine the degree of $\mathcal{NC}$ present in the hidden layers of three different neural network classifiers. Two of these models are popular in the computer vision community, and have been extensively studied and widely adopted: VGG11 \cite{vgg} and ResNet18 \cite{simonyan2014very}. We also train a fully-connected (FC) classification network, MLP6, with network depth of $\ell=6$ and layer width of $d=4096$ for each of its hidden layers. MLP6 serves as a toy model in which to more easily explore $\mathcal{NC}$. In addition to varying network architecture, we also vary the activation functions. Specifically, we explore the effects on $\mathcal{NC}$ of the {\tt ReLU}, {\tt Tanh}, and {\tt LeakyReLU} activation functions. We train these classification neural networks on five popular computer vision classification datasets: MNIST \cite{lecun1998mnist}, CIFAR10 and CIFAR100 \cite{krizhevsky2009learning}, SVHN \cite{netzer2011reading}, and FashionMNIST \cite{xiao2017fashion}. To rebalance the datasets, MNIST and SVHN were subsampled to $N= 5,000$, $N = 4,600$, and $N= 600$ images per class, respectively. We normalize all datasets but do not perform any data augmentation. We use stochastic gradient descent with 0.9 momentum, Mean Square Error Loss (MSE)~\footnote{We use MSE loss rather than Cross Entropy loss as it has been shown to exhibit a greater degree of $\mathcal{NC}$ in classification neural networks while maintaining a greater degree of mathematical clarity \cite{papyan2020prevalence, papyan2019measurements}.}, $\lambda=10^{-5}$ weight decay, and the one-cycle learning rate scheduler for all training \cite{smith2018superconvergence}.

\subsection{Intermediate Layer Analysis}\label{subsec:nc-proper-def}
To assess the extent of $\mathcal{NC}$ in the hidden layers of our classifiers, we perform a step of ``$\mathcal{NC}$ analysis'' at various points during training. This analysis involves freezing the network and passing all the training samples through it. We then collect the network's post-activation representation of each sample after every hidden layer of interest. Specifically, we collect post-activations after each FC layer in MLP6, each convolutional layer in VGG11, and each convolutional layer in ResNet18. We then flatten these post-activation representations into vectors $\bm h^j_{i, c}$, where $j$ is the hidden layer, $i$ is the training sample, and $c$ is the class. We then compute four quantities in each of these hidden layer post-activation vectors to track $\mathcal{NC}$: intra-class variance collapse $(\mathcal{NC}1)$, intra-class norm equality $(\mathcal{NC}2)$, inter-class maximal angular separation $(\mathcal{NC}2)$, and simplification to nearest-class center classifier $(\mathcal{NC}4)$ following the general outline provided in \cite{papyan2020prevalence}. The specifics of these quantities are provided in \Cref{sec:appendixa}. 

\section{Results}
\begin{figure}[th]
    \centering
    \includegraphics[width=.99\linewidth]{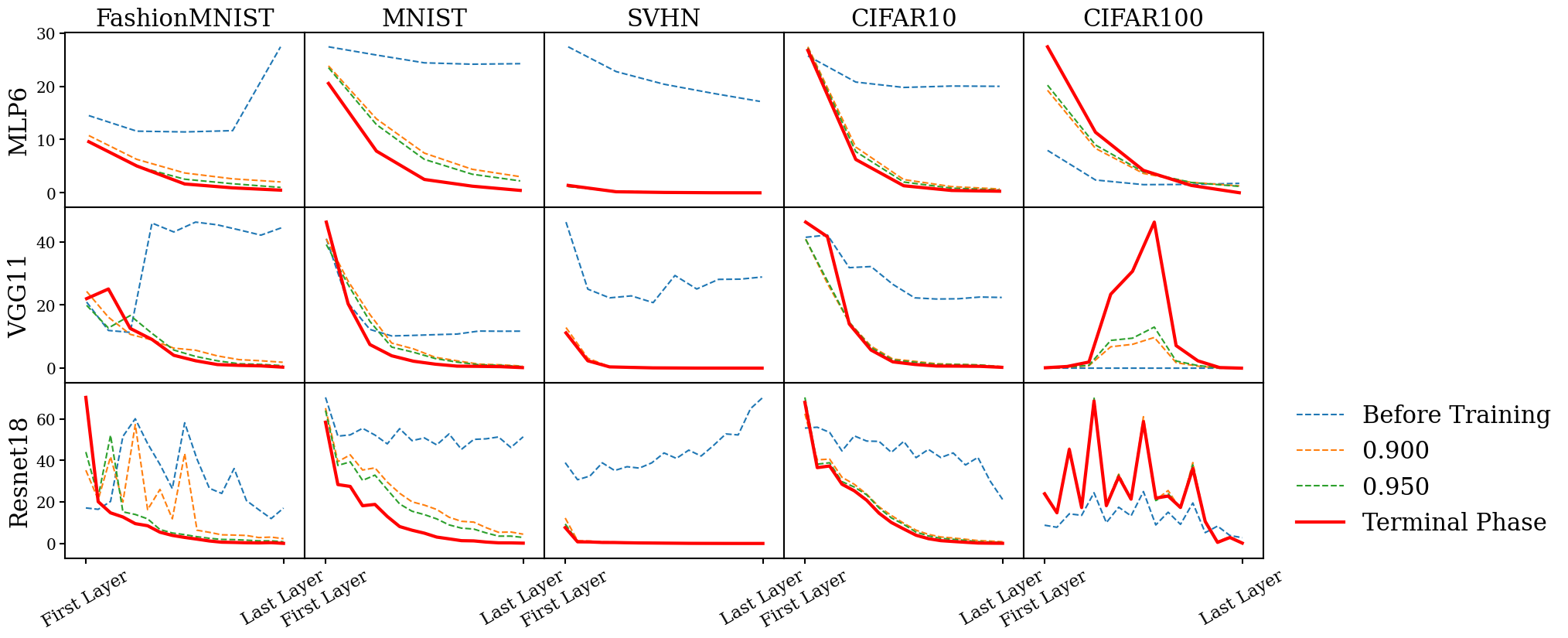}
    \caption{\textbf{Intra-class Variance Collapse ($\mathcal{NC}1$) in the {\tt ReLU} classifiers' intermediate hidden layers.} The results are generated at various points in training, where the blue dotted line indicates $\mathcal{NC}1$ at initialization and the red solid line indicates $\mathcal{NC}1$ after TPT.}
    \label{fig:ReLU_nc1}
\end{figure}

\begin{figure}[th]
    \centering
    \includegraphics[width=.99\linewidth]{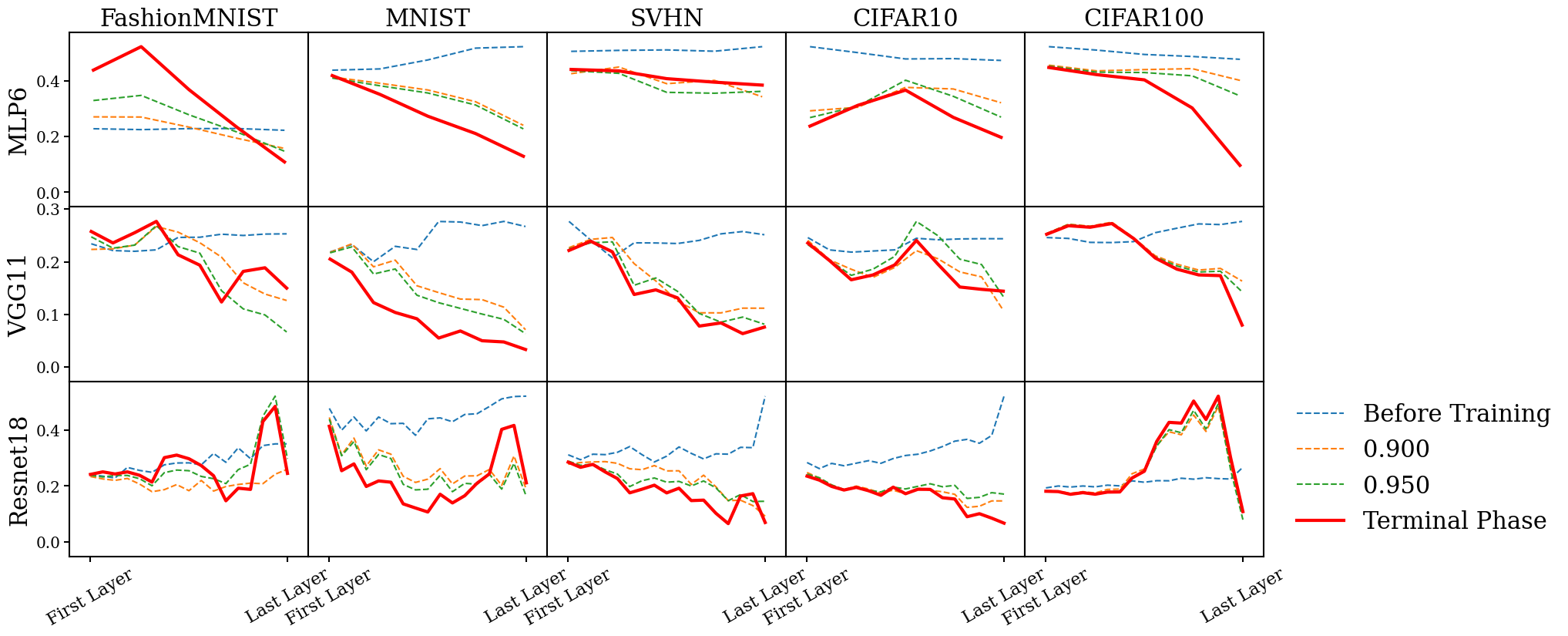}
    \caption{\textbf{Equal Norms ($\mathcal{NC}2$) in the {\tt ReLU} classifiers' intermediate hidden layers.} The results are generated at various points in training, where the blue dotted line indicates $\mathcal{NC}2$ at initialization and the red solid line indicates $\mathcal{NC}2$ after TPT.}
    \label{fig:ReLU_nc2a}
\end{figure}

\begin{figure}[th]
    \centering
    \includegraphics[width=.99\linewidth]{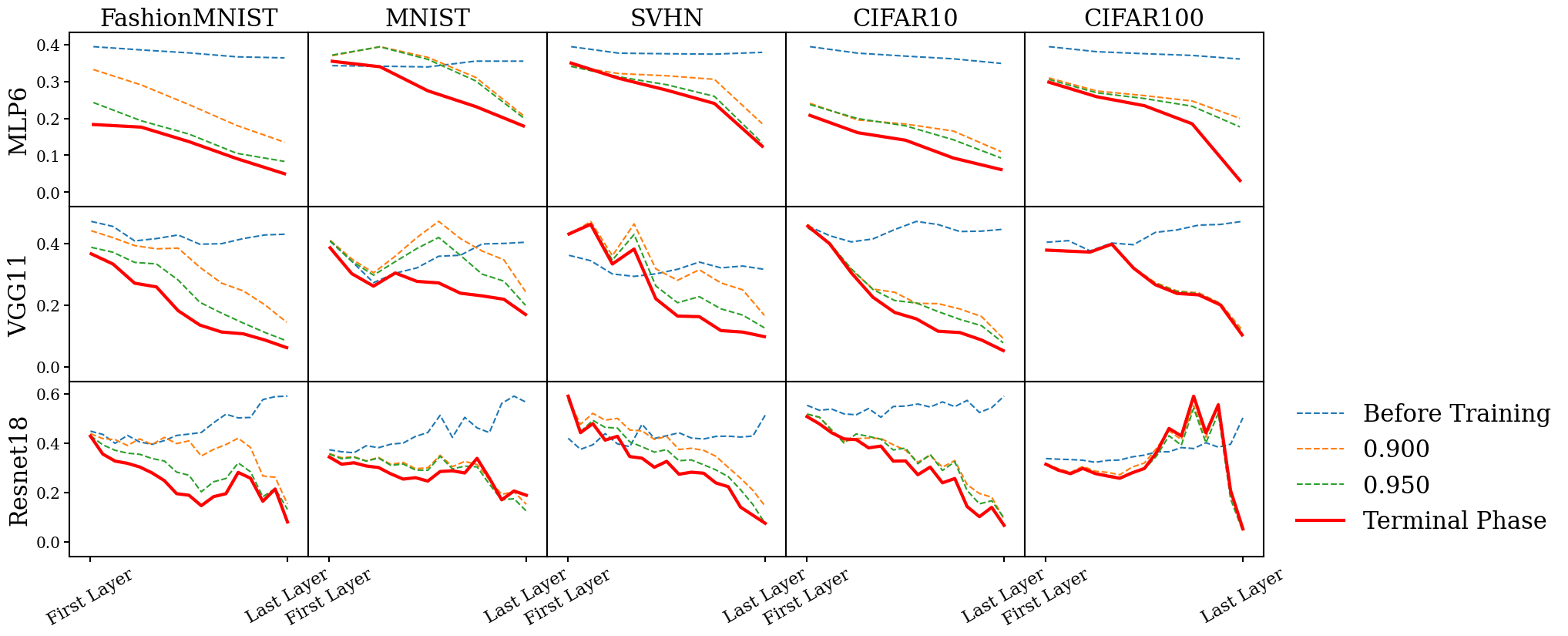}
    \caption{\textbf{Maximal Angles ($\mathcal{NC}2$) in the {\tt ReLU} classifiers' intermediate hidden layers.} The results are generated at various points in training, where the blue dotted line indicates $\mathcal{NC}2$ at initialization and the red solid line indicates $\mathcal{NC}2$ after TPT.}
    \label{fig:ReLU_nc2b}
\end{figure}

\begin{figure}[th]
    \centering
    \includegraphics[width=.99\linewidth]{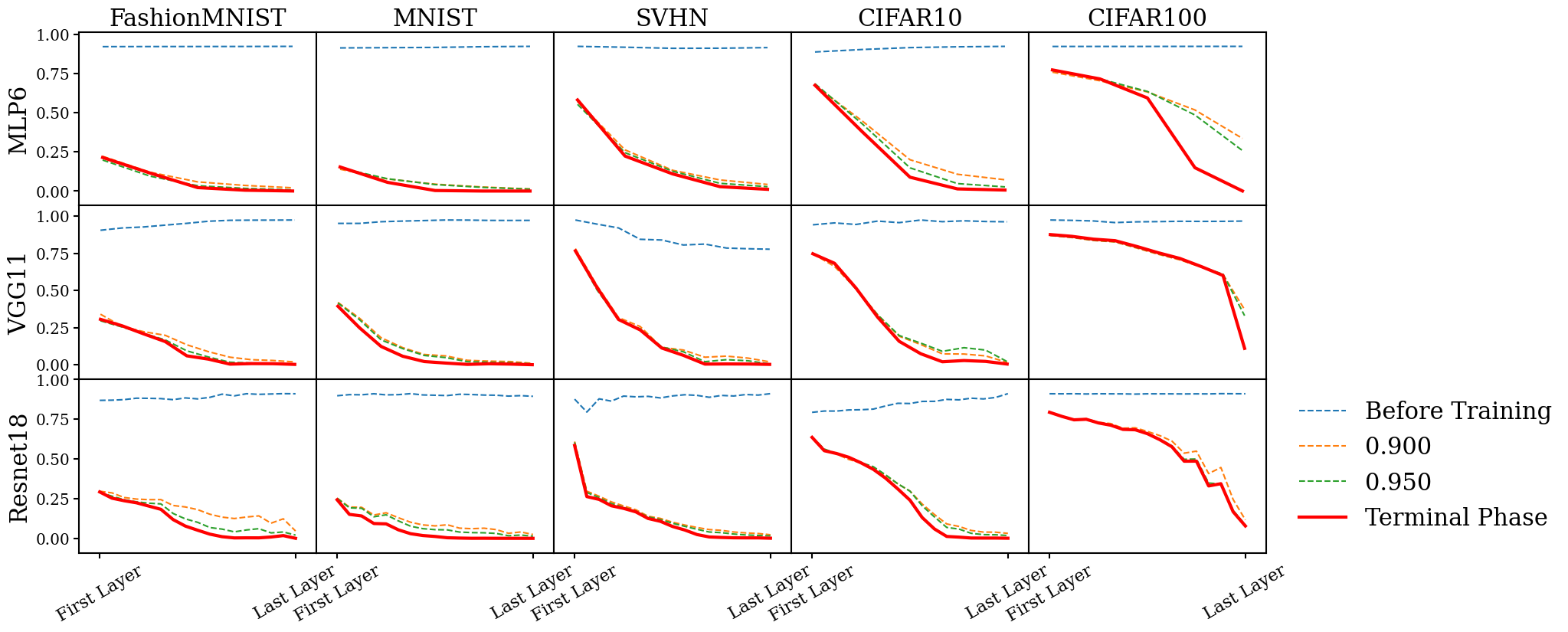}
    \caption{\textbf{Simplification to NCC ($\mathcal{NC}4$) in the {\tt ReLU} classifiers' intermediate hidden layers.} The results are generated at various points in training, where the blue dotted line indicates $\mathcal{NC}4$ at initialization and the red solid line indicates $\mathcal{NC}4$ after TPT.}
    \label{fig:ReLU_nc4}
\end{figure}

We present the results for each of the $\mathcal{NC}$ conditions for MLP6, VGG11, and ResNet18 with {\tt ReLU} activation functions trained to TPT on FashionMNIST, MNIST, SVHN, CIFAR10, and CIFAR100 in \Cref{fig:ReLU_nc1} to \Cref{fig:ReLU_nc4}. 

$\mathcal{NC}1$: The within-class variability decreases linearly with layer depth in the shallower layers of the classification networks, indicating that the earlier fully-connected/convolutional layers are equally effective in clustering samples of the same class. However, in the deeper layers of the networks, the within-class variability plateaus, suggesting that the network has already maximally reduced the variance between same-class samples even when they have only partially propagated through the network. This behavior is observed in most tested network architectures and datasets, except for CIFAR100. Ultimately, the earlier layers of the classifiers primarily group same-class samples together and contribute to the generation of $\mathcal{NC}1$.

$\mathcal{NC}2$: The two phenomena related to the emergence of the simplex ETF structure in the class means, i.e. the convergence of class means to equal norms and to maximal equal angles, also exhibit a somewhat linear relationship between the degree of collapse in any given hidden layer and that layer's depth in the network. However, unlike $\mathcal{NC}1$, the collapse continues to strengthen even in the deeper layers of the network, rather than plateauing after the first few layers; this is more prevalent for the angular separation between different class means than for the similarity between their norms. This phenomenon also persists across most architectures and datasets, and suggests that the network continues to separate classes as it feeds samples forward through its full depth. This makes sense, as features extracted in the shallower layers of the network can be used to learn more and more effective ways of separating different-class samples in deeper layers, leading to an increase in the recorded strength of $\mathcal{NC}2$ over layer depth. 

$\mathcal{NC}4$: The degree of $\mathcal{NC}4$ in any given layer during training seems to be influenced by the presence of $\mathcal{NC}1$ and $\mathcal{NC}2$ in that layer. For the nearest class mean to accurately predict network output, samples need to be close to their class mean ($\mathcal{NC}1$) and class means should be well-separated ($\mathcal{NC}2$). In most experiments, $\mathcal{NC}4$ decreases linearly in the shallower layers and plateaus in the deeper layers. However, the plateauing occurs later than $\mathcal{NC}1$ due to the additional angular separation between class means in these deeper layers, which contributes valuable information for classification. The degree of $\mathcal{NC}4$ in the $j$-th hidden layer indicates how much of the network's classification ability is captured by its $\leq j$ hidden layers. If the mismatch between the nearest neighbor classification in the $j$-th layer and the total network classification is zero, then all of the network's classification ability is already present by the $j$-th layer. For simpler datasets like MNIST and SVHN, the NCC mismatch between shallower layers and the network output reaches zero, while for more complex datasets like CIFAR100, the NCC mismatch only reaches zero in the final layers. This observation aligns with the notion that complex datasets require the deeper networks for complete classification, while simpler datasets can achieve it with the shallower layers; however, it will be important for future studies to observe how this generalizes to test/validation datasets. 

However, despite these general trends, we note that the models trained on CIFAR100 exhibit a number of unique behaviors not consistent with our broader observations on $\mathcal{NC}$, most striking of which is no significant decrease in within-class variability over training. Instead, the data seems largely noisy for this $\mathcal{NC}$ condition. This merits future investigation across other challenging datasets such as TinyImageNet, as well as more exploratory analysis. 

In addition to the experiments performed on the classification networks with {\tt ReLU} activation functions above, we also perform the same set of experiments on {\tt Tanh} and {\tt LeakyReLU} classifiers in \Cref{sec:tanh} and \Cref{sec:leakyReLU} respectively. These experiments largely demonstrate the same characteristics as the {\tt ReLU} experiments.

\section{Conclusions}
Our work demonstrates that the \textit{Neural Collapse} ($\mathcal{NC}$) phenomenon occurs throughout most of the hidden layers of classification neural networks trained through the terminal phase of training (TPT). We demonstrate these findings across a variety of settings, including varying network architectures, classification datasets, and network activations functions, and find that the degree of $\mathcal{NC}$ present in any hidden layer is typically correlated with the depth of that layer in the network. In particular, we make the following specific observations: (1) almost all of the reduction in intra-class variance in the samples occurs in the shallower layers of the classification networks, (2) the angular separation between class means is increased consistently as samples propagate through the entire network, and (3) simpler datasets require only the shallower layers of the networks to fully learn them, whereas more difficult ones require the entire network. Ultimately, these results provide a granular view of the structural propagation of features through classification networks. In future work, it will be important to analyze how these results generalize to held-out validation data. For example, do our observations of $\mathcal{NC}4$ serve as a proxy for the $\leq j$ network's ability to classify data extend to validation data? Moreover, is there a broader relationship between $\mathcal{NC}$ and network generalization/over-training?

\section{Contributions}
\textbf{Parker}: Initiated problem concept; led experiments and analysis design; collaborated on computational work; collaborated on results analysis; wrote the paper. \textbf{Onal}: Collaborated on experiments and analysis design; collaborated on computational work; collaborated on results analysis. \textbf{Stengel}: Collaborated on background research and initial experiment design; collaborated on computational work. \textbf{Intrater}: Collaborated on problem concept and background research. 

\section{Acknowledgements}
We would like to thank Professor Boris Hanin for his generous guidance and support in the completion of this paper as well as for his insightful and exciting class, Deep Learning Theory, which ultimately led to the creation of this project.

\printbibliography

\clearpage
\appendix

\section{Intermediate Layer Analysis}\label{sec:appendixa}
In this section, we provide greater details on how the various $\mathcal{NC}$ conditions are computed from the flattened $\bm h^j$ post-activation vectors after they have been randomly sampled (if they have been randomly sampled at all). Specifically, from these post-activation vectors, we first compute four preliminary quantities for each layer $j$, which we will use later to calculate the $\mathcal{NC}$ metrics:

\begin{enumerate}
    \item $j$-th layer global mean: $\bm \mu_{G}^j = \operatorname{Ave}^j_{i,c} \bm h_{i,c}^j$,
    \item $j$-th layer class means: $\bm \mu_{c}^j = \operatorname{Ave}^j_{i} \bm h_{i,c}^j$,
    \item $j$-th layer within-class covariance: 
    $\Sigma_{W}^j = \operatorname{Ave}^j_{i} (\bm h_{i,c}^j - \bm\mu_{c}^j)(\bm h_{i,c}^j - \bm\mu_{c}^j)^T \in \mathbb{R}^{p_j\times p_j}$,
    \item $j$-th layer between-class covariance: 
    $\Sigma_{B}^j = \operatorname{Ave}_{c}^j (\bm\mu_{c}^j - \bm\mu_{G}^j)(\bm\mu_{c}^j - \bm\mu_{G}^j)^T \in \mathbb{R}^{p_j\times p_j}$,
\end{enumerate}

From these quantities, we compute the following for each of the hidden layers across the entire training set, following the general outline provided by \cite{papyan2020prevalence}. Importantly, we do not track $\mathcal{NC}3$, as it does not yield any meaningfully intuitive insights in the context of intermediate layer analysis:
\\

\noindent $(\mathcal{NC}1)$ \textbf{Intra-Class Variance Collapse:} 
\[\operatorname{Tr}(\bm\Sigma_B^{j +} \bm\Sigma_W^j/C).\]
Here $+$ denotes the pseudoinverse calculated with singular value decomposition (SVD). Ignoring the first term, the variation becomes negligible as the weights approach their class means; thus, the vanishing of this quantity implies the vanishing of intra-class variance in the intermediate activation maps. This quantity is a more precise version of the one introduced in \cite{papyan2020prevalence}.
\\

\noindent $(\mathcal{NC}2)$ \textbf{Convergence to Simplex ETF:} The first quantity, "Equal Norms", is
\[    \operatorname{std}^j_c(\|\bm\mu_c^j - \bm\mu_G^j\|_2)/\operatorname{avg}^j_c(\|\bm\mu_c^j-\bm\mu_G^j\|_2).\]
As this quantity converges to zero, the coefficients of variation of class means similarly vanish, implying that the class-means become equinormal. The second quantity, "Maximal Angles", is
\[\operatorname{avg}_{c\neq c'}\left(\left| \frac{\langle\bm\mu_c^j-\mu_G^j \bm\mu_{c'}^j-\bm\mu_G^j\rangle}{\|\bm\mu_c^j-\bm\mu_G^j\|_2\|\bm\mu_{c'}^j-\bm\mu_G^j\|_2} + \frac{1}{C-1}\right|\right).\]
 The first component in the average denotes $\cos_\mu(c^j, c'^j)$, i.e. the angle between the average activation map of class $c$ and that of class $c'$ in layer $j$. Thus this represents the average difference between these angles and $\frac{-1}{C-1}$, and therefore the convergence to zero corresponds to maximum separation for globally centered vectors. Together, these terms represent convergence of the class means to a Simplex ETF.
\\
 
\noindent $(\mathcal{NC}4)$ \textbf{Simplification to Nearest Class-Center:}
\[1 - \operatorname{avg}_{i,c}\mathbb{1}\{f_\theta(x_{i,c}) = \arg\min_{c'}\| \bm h^j_{i,c}-\bm \mu_{c'}^j\|_2\}\]

Here $1 \{.\}$ is the indicator function. This quantity represents the proportion of times that the entire network $f_\theta$ disagrees with the result that would have been obtained by simply taking the nearest class mean for an activation map in the $j$th layer. As this term goes to zero, it implies that the classifier's behavior simplifies to a nearest class-mean decision rule. Importantly, as this quantity vanishes in intermediate layers, it implies that the features necessary for the classifier to correctly classify 

\section{{\tt Tanh} Activations}\label{sec:tanh}
We present the results for each of the $\mathcal{NC}$ conditions for the various architectures with {\tt Tanh} activation functions trained to TPT on the various datasets in \Cref{fig:tanh_nc1} to \Cref{fig:tanh_nc4}. 

\begin{figure}[H]
    \centering
    \includegraphics[width=.99\linewidth]{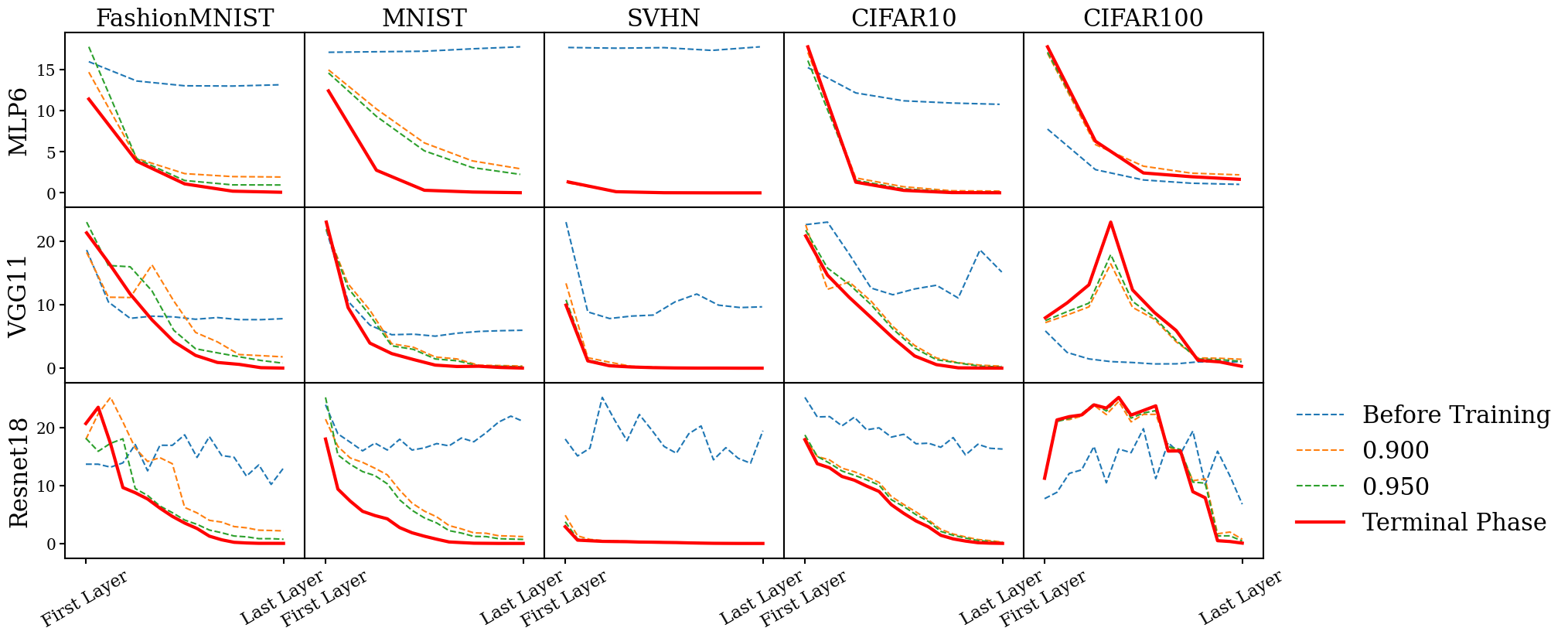}
    \caption{\textbf{Intra-class Variance Collapse ($\mathcal{NC}1$) in the {\tt Tanh} classifiers' intermediate hidden layers.} The results are generated at various points in training, where the blue dotted line indicates $\mathcal{NC}1$ at initialization and the red solid line indicates $\mathcal{NC}1$ after TPT.}
    \label{fig:tanh_nc1}
\end{figure}

\begin{figure}[H]
    \centering
    \includegraphics[width=.99\linewidth]{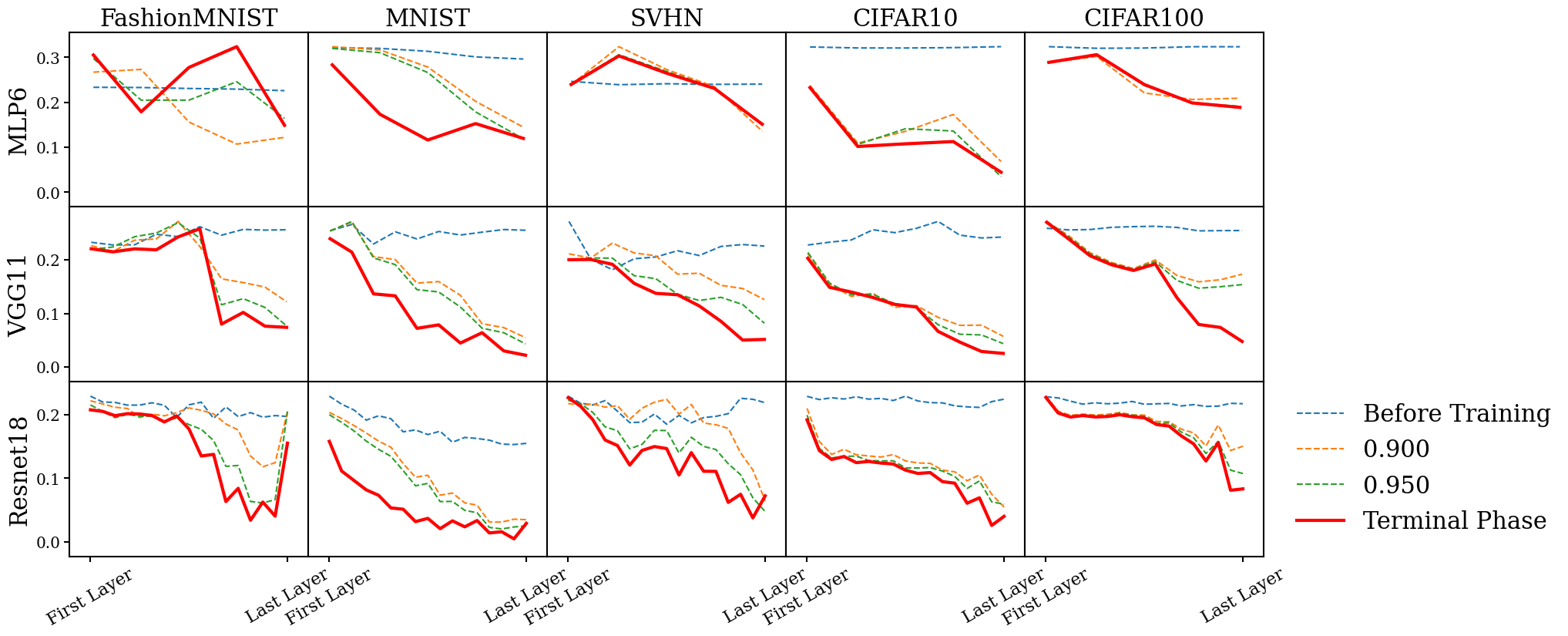}
    \caption{\textbf{Equal Norms ($\mathcal{NC}2$) in the {\tt Tanh} classifiers' intermediate hidden layers.} The results are generated at various points in training, where the blue dotted line indicates $\mathcal{NC}2$ at initialization and the red solid line indicates $\mathcal{NC}2$ after TPT.}
    \label{fig:tanh_nc2a}
\end{figure}

\begin{figure}[H]
    \centering
    \includegraphics[width=.99\linewidth]{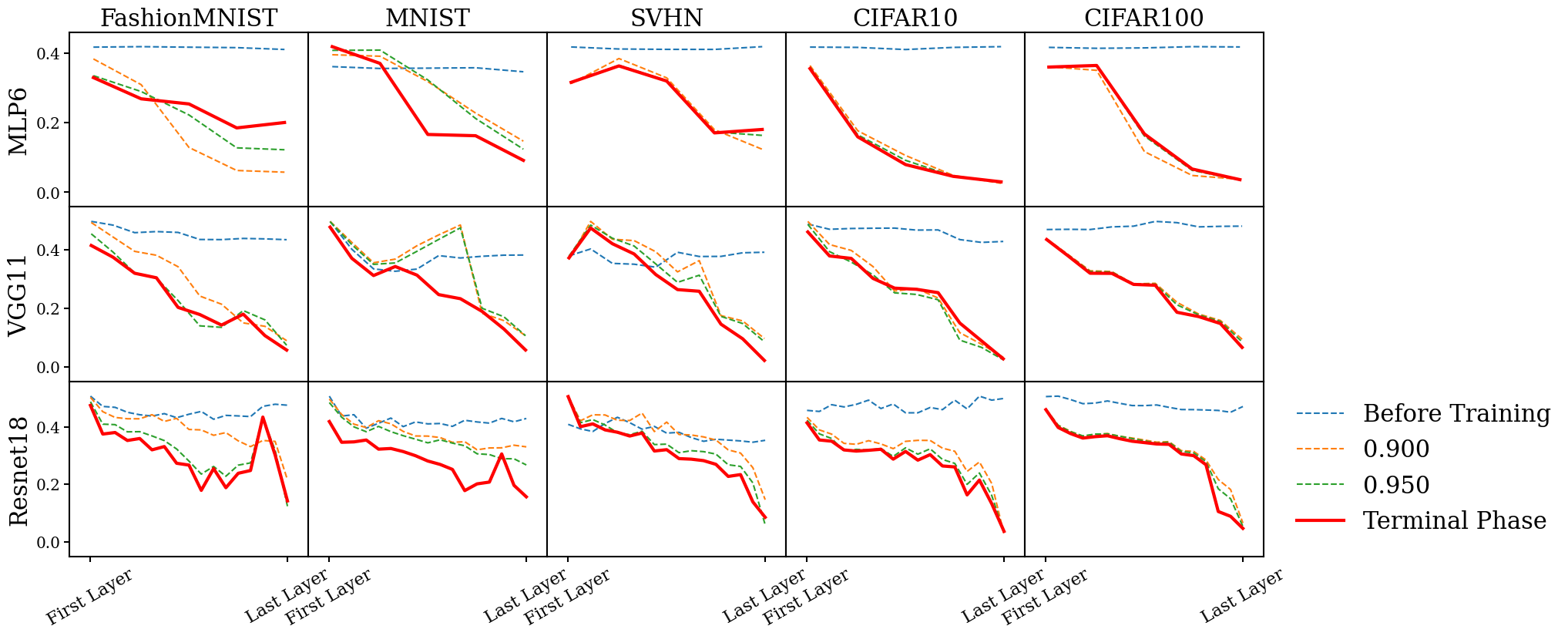}
    \caption{\textbf{Maximal Angles ($\mathcal{NC}2$) in the {\tt Tanh} classifiers' intermediate hidden layers.} The results are generated at various points in training, where the blue dotted line indicates $\mathcal{NC}2$ at initialization and the red solid line indicates $\mathcal{NC}2$ after TPT.}
    \label{fig:tanh_nc2b}
\end{figure}

\begin{figure}[H]
    \centering
    \includegraphics[width=.99\linewidth]{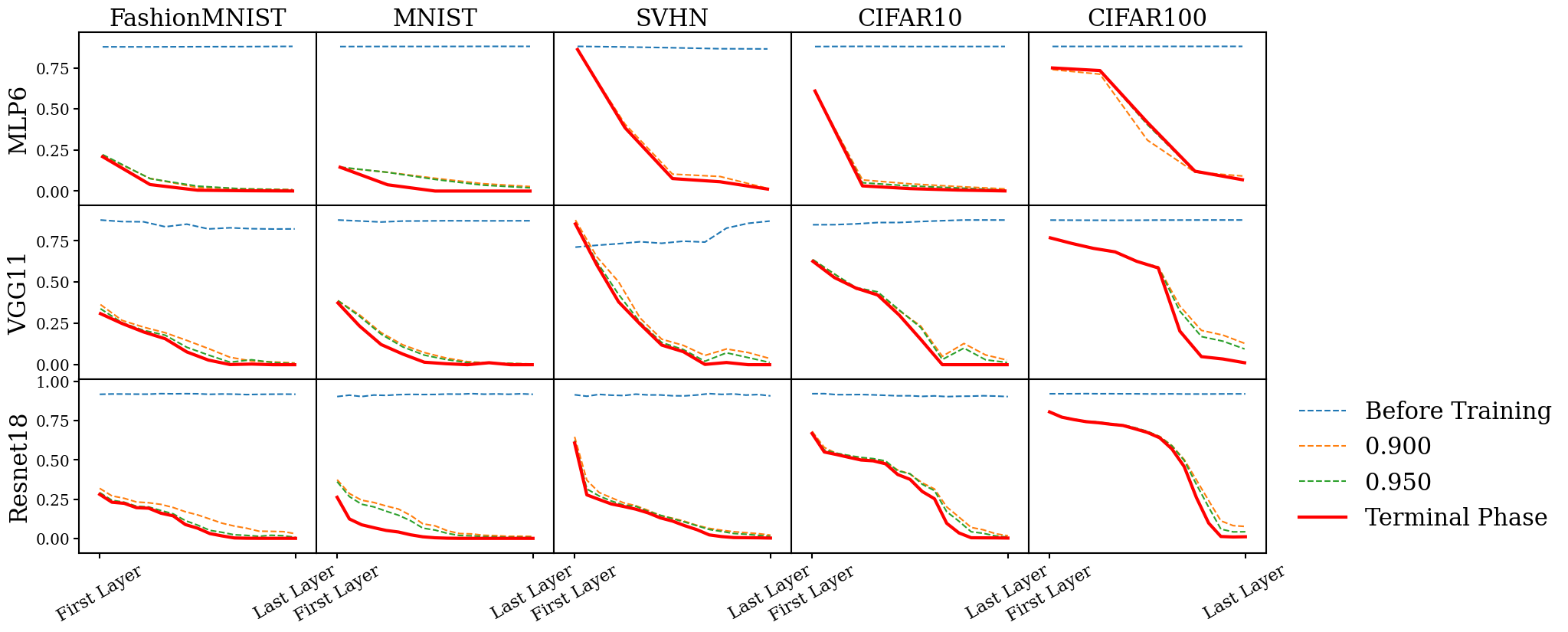}
    \caption{\textbf{Simplification to NCC ($\mathcal{NC}4$) in the {\tt Tanh} classifiers' intermediate hidden layers.} The results are generated at various points in training, where the blue dotted line indicates $\mathcal{NC}4$ at initialization and the red solid line indicates $\mathcal{NC}4$ after TPT.}
    \label{fig:tanh_nc4}
\end{figure}

\section{{\tt LeakyReLU} Activations}\label{sec:leakyReLU}
We present the results for each of the $\mathcal{NC}$ conditions for the various architectures with {\tt LeakyReLU} activation functions trained to TPT on the various datasets in \Cref{fig:tanh_nc1} to \Cref{fig:tanh_nc4}.

\begin{figure}[H]
    \centering
    \includegraphics[width=.99\linewidth]{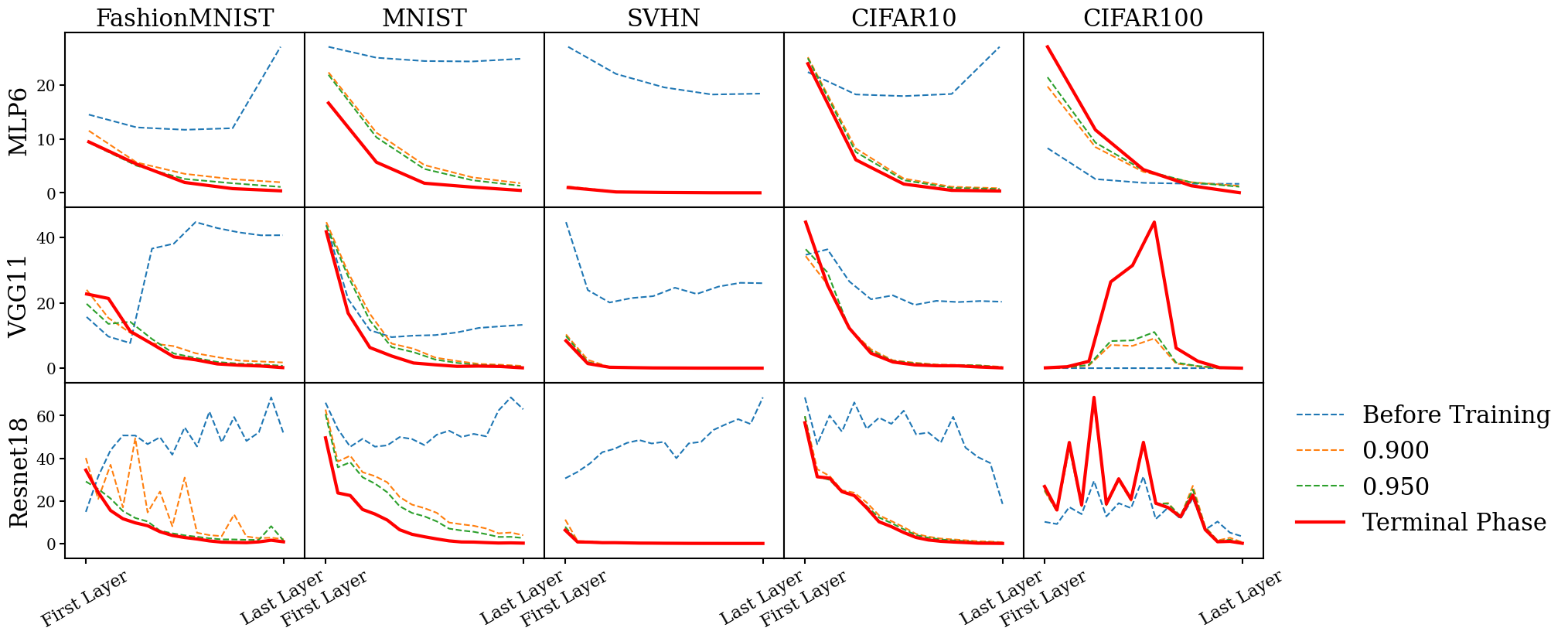}
    \caption{\textbf{Intra-class Variance Collapse ($\mathcal{NC}1$) in the {\tt LeakyReLU} classifiers' intermediate hidden layers.} The results are generated at various points in training, where the blue dotted line indicates $\mathcal{NC}1$ at initialization and the red solid line indicates $\mathcal{NC}1$ after TPT.}
    \label{fig:tanh_nc1}
\end{figure}

\begin{figure}[H]
    \centering
    \includegraphics[width=.99\linewidth]{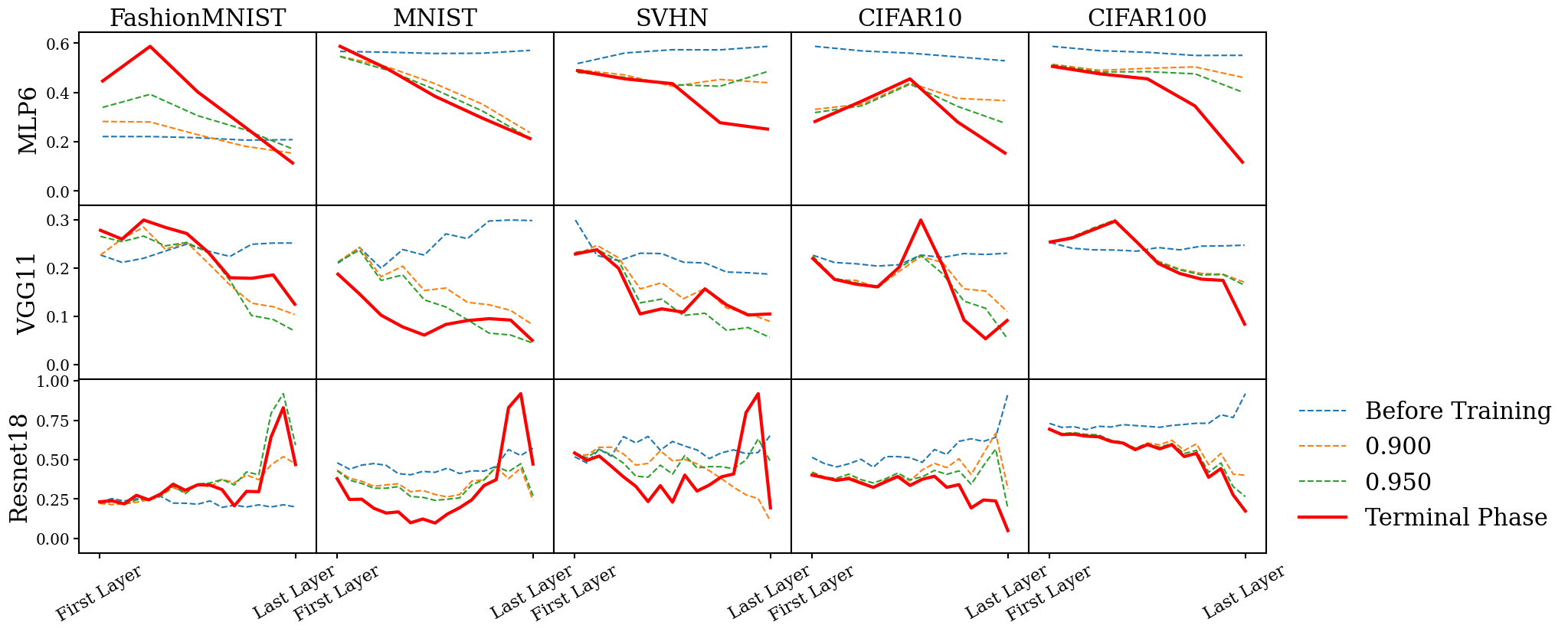}
    \caption{\textbf{Equal Norms ($\mathcal{NC}2$) in the {\tt LeakyReLU} classifiers' intermediate hidden layers.} The results are generated at various points in training, where the blue dotted line indicates $\mathcal{NC}2$ at initialization and the red solid line indicates $\mathcal{NC}2$ after TPT.}
    \label{fig:tanh_nc2a}
\end{figure}

\begin{figure}[H]
    \centering
    \includegraphics[width=.99\linewidth]{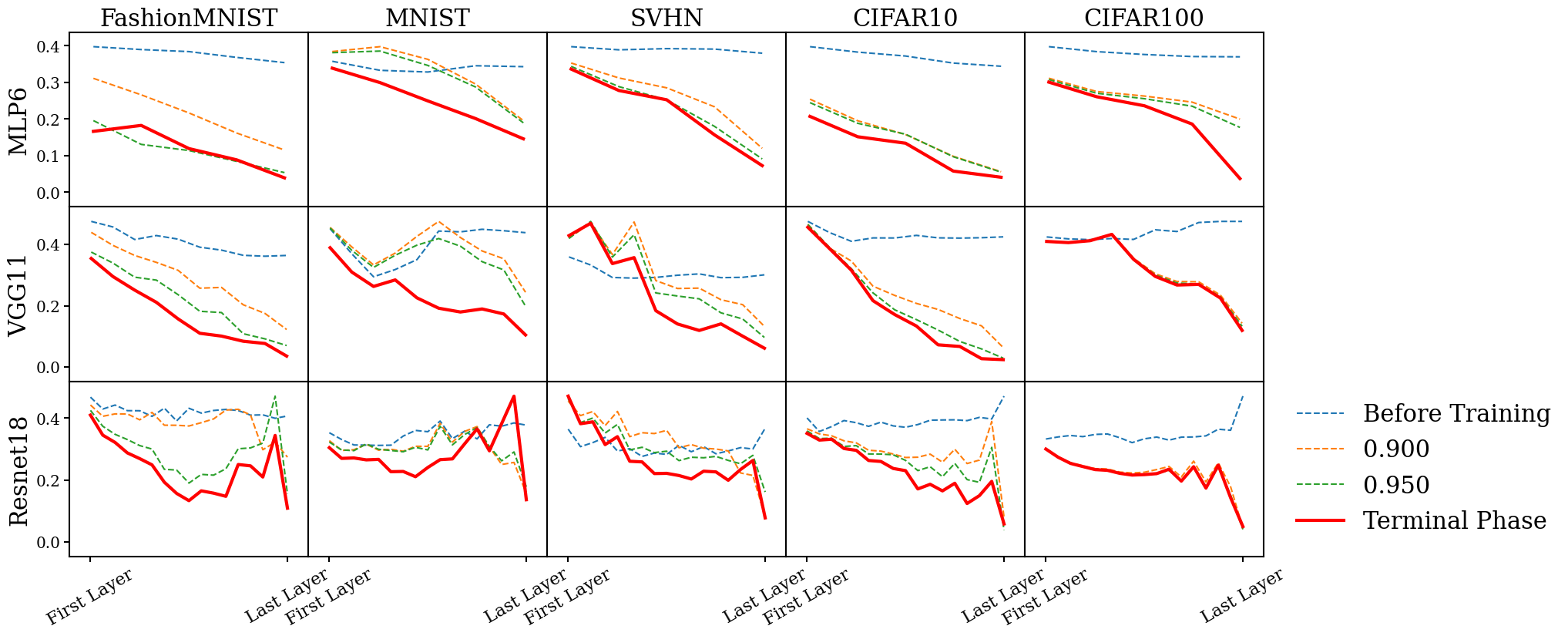}
    \caption{\textbf{Maximal Angles ($\mathcal{NC}2$) in the {\tt LeakyReLU} classifiers' intermediate hidden layers.} The results are generated at various points in training, where the blue dotted line indicates $\mathcal{NC}2$ at initialization and the red solid line indicates $\mathcal{NC}2$ after TPT.}
    \label{fig:tanh_nc2b}
\end{figure}

\begin{figure}[H]
    \centering
    \includegraphics[width=.99\linewidth]{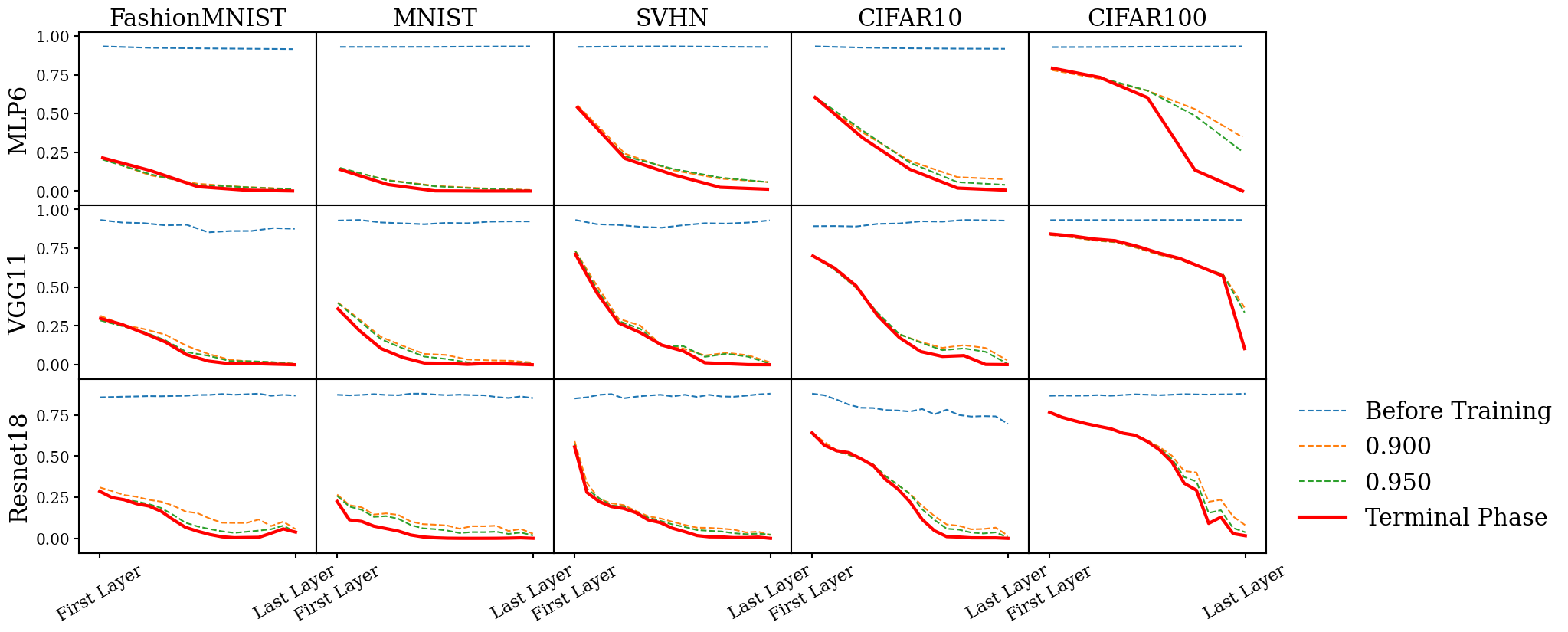}
    \caption{\textbf{Simplification to NCC ($\mathcal{NC}4$) in the {\tt LeakyReLU} classifiers' intermediate hidden layers.} The results are generated at various points in training, where the blue dotted line indicates $\mathcal{NC}4$ at initialization and the red solid line indicates $\mathcal{NC}4$ after TPT.}
    \label{fig:tanh_nc4}
\end{figure}

\end{document}